\documentclass{article}

\PassOptionsToPackage{numbers, compress}{natbib}

\usepackage[preprint]{neurips_data_2022}

\usepackage[utf8]{inputenc} 
\usepackage[T1]{fontenc}    
\usepackage{microtype}
\usepackage{times}
\usepackage{graphicx}
\usepackage{amsmath,amssymb,mathbbol}
\usepackage{algorithmic}
\usepackage[linesnumbered,ruled,vlined]{algorithm2e}
\usepackage{acronym}
\usepackage{enumitem}
\usepackage[pagebackref,colorlinks,breaklinks,citecolor=gray]{hyperref}
\usepackage{balance}
\usepackage{xspace}
\usepackage{setspace}
\usepackage[skip=3pt,font=small]{subcaption}
\usepackage[skip=3pt,font=small]{caption}
\usepackage[dvipsnames]{xcolor}
\usepackage[capitalise]{cleveref}
\usepackage{booktabs,tabularx,colortbl,multirow,array,makecell}
\usepackage{wrapfig}
\usepackage{siunitx} 
\usepackage{soul} 

\makeatletter
\DeclareRobustCommand\onedot{\futurelet\@let@token\@onedot}
\def\@onedot{\ifx\@let@token.\else.\null\fi\xspace}
\def\eg{\emph{e.g}\onedot} 

\def\ie{\emph{i.e}\onedot}

\def\vs{\emph{vs}\onedot}

\makeatother

\makeatletter
\renewcommand{\paragraph}{%
  \@startsection{paragraph}{4}%
  {\z@}{0ex \@plus 0ex \@minus 0ex}{-1em}%
  {\hskip\parindent\normalfont\normalsize\bfseries}%
}
\makeatother

\graphicspath{{figures/}}

\crefname{algorithm}{Alg.}{Algs.}
\Crefname{algocf}{Algorithm}{Algorithms}
\crefname{section}{Sec.}{Secs.}
\Crefname{section}{Section}{Sections}
\crefname{table}{Tab.}{Tabs.}
\Crefname{table}{Table}{Tables}
\crefname{figure}{Fig.}{Fig.}
\Crefname{figure}{Figure}{Figure}

\definecolor{gblue}{HTML}{4285F4}
\definecolor{gred}{HTML}{DB4437}
\definecolor{ggreen}{HTML}{0F9D58}

\definecolor{mygray}{gray}{.92}

\acrodef{bm}[G-VUE]{General-purpose Visual Understanding Evaluation}
\acrodef{nlp}[NLP]{Natural Language Processing}

\newcommand{\comment}[1]{}

\acrodef{glue}[GLUE]{General Language Understanding Evaluation}
\acrodef{mse}[MSE]{mean-squared error}
\acrodef{miou}[mIoU]{mean Intersection over Union}

\title{\textit{Perceive}, \textit{Ground}, \textit{Reason}, and \textit{Act}: A Benchmark for General-purpose Visual Representation}

\author{
    \textbf{Jiangyong Huang}$^{1,3,*}$ \\
    \And \textbf{William Yicheng Zhu}$^{1,*}$ \\
    \And \textbf{Baoxiong Jia}$^{1,2}$ \\
    \AND \textbf{Zan Wang}$^{1,4}$ \\
    \And \textbf{Xiaojian Ma}$^{1,2}$ \\
    \And \textbf{Qing Li}$^1$ \\
    \And \textbf{Siyuan Huang}$^{1}$ \\
    \AND
    $^1$ \normalfont{Beijing Institute for General Artificial Intelligence}\\
    $^2$ University of California, Los Angeles \\
    $^3$ Peking University\\
    $^4$ Beijing Institute of Technology \\
}
\begin{document}

\maketitle
\let\thefootnote\relax\footnotetext{* indicates equal contribution.}

\begin{abstract}
Current computer vision models, unlike the human visual system, cannot yet achieve general-purpose visual understanding. Existing efforts to create a general vision model are limited in the scope of assessed tasks and offer no overarching framework to perform them holistically. We present a new comprehensive benchmark, General-purpose Visual Understanding Evaluation (G-VUE), covering the full spectrum of visual cognitive abilities with four functional domains — \textit{Perceive}, \textit{Ground}, \textit{Reason}, and \textit{Act}. The four domains are embodied in 11 carefully curated tasks, from 3D reconstruction to visual reasoning and manipulation. Along with the benchmark, we provide a general encoder-decoder framework to allow for the evaluation of arbitrary visual representation on all 11 tasks. We evaluate various pre-trained visual representations with our framework and observe that (1) Transformer-based visual backbone generally outperforms CNN-based backbone on G-VUE, (2) visual representations from vision-language pre-training are superior to those with vision-only pre-training across visual tasks. With G-VUE, we provide a holistic evaluation standard to motivate research toward building general-purpose visual systems via obtaining more general-purpose visual representations.
\end{abstract}


\section{Introduction}
The long-term goal of machine vision~\citep{marr1982vision,barrow1981computational,ikeuchi1996task,marr2010vision} is to build a general-purpose vision system that can perceive, understand, and react to visual inputs from unconstrained environments. The term \textit{general-purpose} can be best understood through observing our own visual systems, which support various complex higher-order visual tasks, from edge detection and object recognition to visual navigation and manipulation, all while rooted in a common brain region that produces core visual representations.

On the other hand, even though the field of computer vision has been thriving with models that solve complex tasks increasingly well over the last decade \citep{he2016deep,he2017mask,ronneberger2015u,dosovitskiy2020image}, there is still a considerable gap between vision models and the aforementioned human visual systems. In particular, current vision models are mostly task-specific and often contain specialized components or architectures designed for a specific setting. They are also hindered by diverse input-output formats that vary from task to task. Recent works in self-supervised learning \citep{oord2018representation,he2020momentum,chen2020simple,grill2020bootstrap,he2021masked,huang2021spatio,ma2022relvit} explore more general visual representations by learning from enormously-sized visual domains. Advances in vision-language modeling \citep{lu202012,hu2021unit,cho2021unifying,gupta2021towards,kamath2021mdetr,wang2022unifying} also seek to unify the vision tasks with a task-agnostic model architecture. However, these models are evaluated solely on traditional tasks like classification, detection, or vision-language understanding. It is still unclear whether current computer vision models are general-purpose like the human visual systems, as there is no overarching benchmark that covers all visual tasks holistically, from low-level perception to high-level reasoning and acting. 

\begin{wrapfigure}{R}{0.48\linewidth}
\vspace{-20px}
  \begin{center}
    \includegraphics[width=\linewidth]{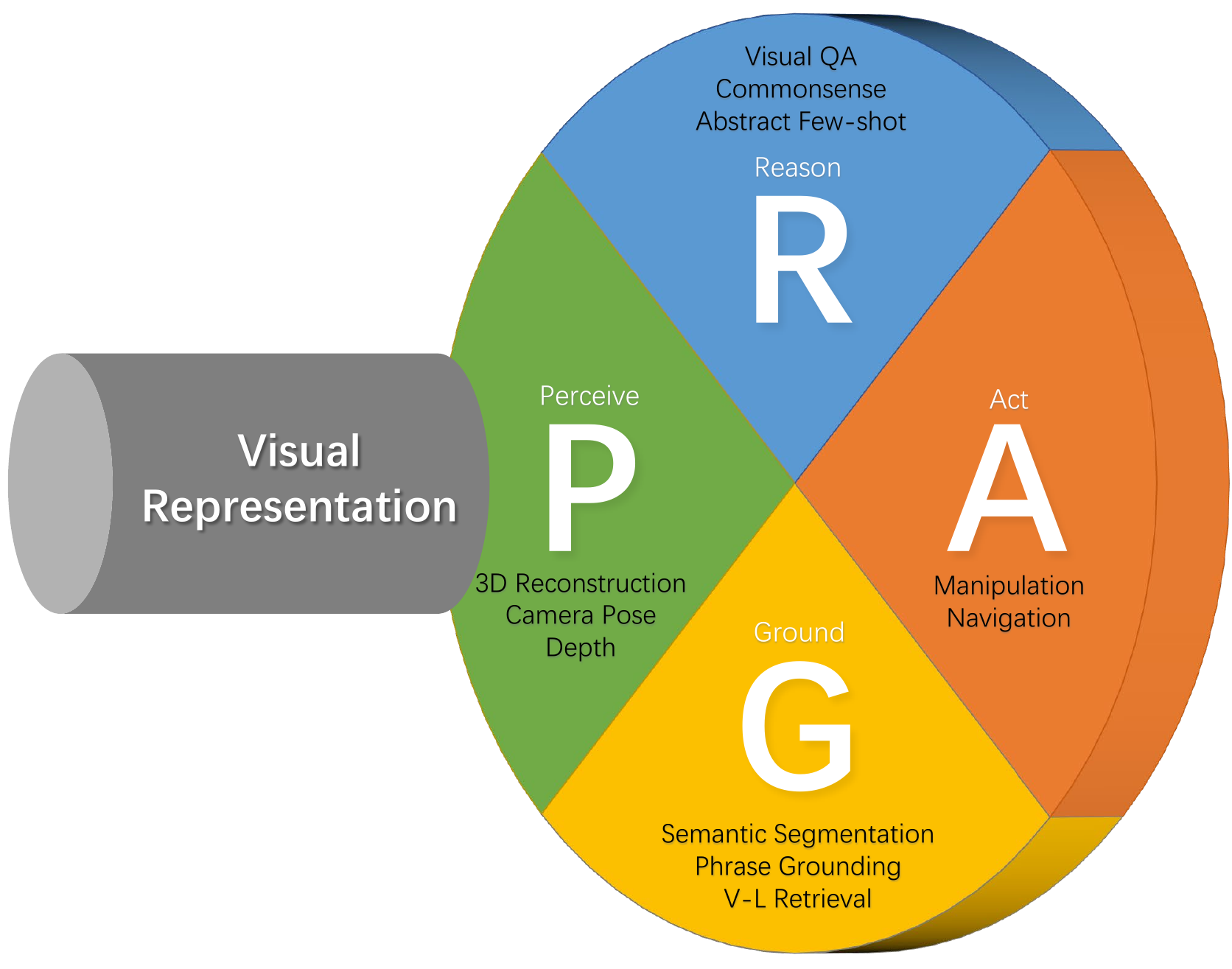}
  \end{center}
  \caption{An overview of G-VUE. The key idea of G-VUE is to evaluate visual representation in a general-purpose standard, where P (\textit{Perceive}), G (\textit{Ground}), R (\textit{Reason}), and A (\textit{Act}) represent four functional domains for a general vision system.}
  \label{fig:teaser}
\vspace{-9px}
\end{wrapfigure}

To facilitate the research toward the general-purpose vision, we present \ac{bm} benchmark. We carefully curate 11 tasks from four functional domains that visual systems should support —\textit{Perceive}, \textit{Ground}, \textit{Reason}, and \textit{Act}— ordered by their cognitive complexity. These four domains cover the full spectrum of human visual tasks and extend far beyond the standard set of visual tasks like detection or V-L grounding discussed in previous models or benchmarks. Specifically, \textit{Perceive} tests a model's geometry understanding. \textit{Ground} examines a model's acquisition of visual semantics. \textit{Reason} probes a model's capacity for logical deduction and common sense reasoning. \textit{Act} investigates a model's ability for planning and decision-making by learning visual policies. \cref{fig:teaser} illustrates our benchmark and \cref{tab:benchmark} shows the task details. Principles and details about task and dataset selection are presented in \cref{sec:gvue} and \cref{sec:design}. \ac{bm} provides a test-bed for a general-purpose vision system and allows for fair comparisons between different visual representations over a full spectrum of visual tasks. 

The diverse visual tasks in \ac{bm} require the model to navigate in a wide range of input and output formats, including image, text, bounding box, dense map, and policy. There is no existing model that can accomplish all these tasks. It is quite different from the \ac{nlp} field, where most tasks \citep{wang2018glue,wang2019superglue} can be accomplished using the same sequence-to-sequence model based on RNN \citep{mikolov2010recurrent}, LSTM \citep{hochreiter1997long}, or Transformer \citep{vaswani2017attention}. Therefore, we carefully equip the visual tasks in \ac{bm} with a general encoder-decoder framework to mitigate the gaps among various visual tasks and accommodate arbitrary visual encoder, which makes it possible to evaluate any visual representation on these 11 tasks. \cref{sec:model} further describes the principles and details of framework design. Notably, although such a framework facilitates handling tasks across various domains, it is not designed for massive multi-task training but for a flexible and fair evaluation of various existing visual representations on G-VUE.


To understand the challenges posed by \ac{bm} and the limitations of current models, we evaluate the most representative visual representations obtained with various architectures (\eg, ResNet vs. ViT) and pre-training mechanisms (\eg, supervised vs. self-supervised). The experimental results show that (1) Transformer-based architectures generally beat CNN-based architectures, and (2) visual representations pre-trained on massive image-text data (\eg, CLIP) significantly outperform traditional ImageNet pre-training across visual tasks.

We make the following contributions: (1) we present a general-purpose vision benchmark \ac{bm} that consists of 11 visual tasks covering diverse visual skills; (2) we introduce an encoder-decoder framework for \ac{bm} that allows for the evaluation of arbitrary visual representations; (3) we examine the performance of representative visual representations on \ac{bm} and present detailed analyses. Additionally, we construct an evaluation platform
and a leaderboard for \ac{bm} that will be made publicly available.

\section{Related Work}
\subsection{Visual Pre-training}
Large-scale visual pre-training has been viewed as a potential way to obtain a general visual representation that adapts well to any downstream tasks. There have been successful attempts at learning robust visual features from discriminative \citep{oord2018representation,he2020momentum,chen2020simple,grill2020bootstrap,chen2021empirical} and generative \citep{he2021masked} self-supervised learning and vision-language alignment learning \citep{radford2021learning}. These models use various visual backbones to learn from diverse visual domains, including images \citep{deng2009imagenet}, egocentric videos \citep{grauman2021ego4d}, and image-text pairs \citep{radford2021learning}. Their differences in learning mechanism, visual backbone and source learning domain affect their transferability to downstream tasks. Although these models have been extensively evaluated in existing tasks such as image classification and object detection, it remains unclear how they can be used to achieve a broader domain of visual tasks toward general-purpose vision. One essential goal of \ac{bm} is to evaluate these models and provide tools for seeking more generalizable pre-trained models.

\subsection{General-purpose Vision}
\paragraph{Models.} The pursuit of general-purpose vision emerges from the early stage of computer vision. In the last 70s, \citet{marr2010vision} introduced a fixed layered representation and unified visual processing for general visual tasks. Despite suffering from its inflexibility to adapt to visual tasks, it provides a paradigm for general visual perception. Recently, progresses in deep learning have driven the vision model towards unified architectures \citep{lu202012,hu2021unit,cho2021unifying,gupta2021towards,kamath2021mdetr,wang2022unifying, lu2022unified} that are agnostic to the input and output formats. Among them, VL-T5 \citep{cho2021unifying} and GPV-1 \citep{gupta2021towards} unify the visual tasks through vision-language models, OFA \citep{wang2022unifying} and UIO \citep{lu2022unified} utilize tokenized representation to model various vision-language tasks jointly. Although these methods tackle a larger set of visual tasks, they are still limited to the conventional vision-language domain, lacking the general capabilities for low-level geometric understanding and high-level reasoning and acting.

\paragraph{Benchmarks.}
The general understanding benchmarks have become a standard in \ac{nlp} \citep{wang2018glue,wang2019superglue} for testing pre-trained language models. However, few benchmarks are designed for general vision understanding, mainly due to the enormous complexity of diverse visual tasks. Among these benchmarks, \citet{zhai2019visual} and \citet{shao2021intern} focus on the transferability of visual representation, which perform domain adaptation by exposing pre-trained visual models to a small fraction of tuning data. GRIT~\citep{gupta2022grit} emphasizes both generality and robustness of visual representation in visual perception and grounding tasks. \citet{parisi2022unsurprising} focuses on the evaluation of pre-trained visual representation on policy learning and explores how ingredients of pre-training affect the learning of control policy. While each of the previous benchmarks only addresses a narrow aspect of human visual function, \ac{bm} provides holistic coverage of visual tasks spanning the functional domains of \textit{Perceive}, \textit{Ground}, \textit{Reason}, and \textit{Act}. With the equipped general encoder-decoder framework, \ac{bm} serves as a platform for evaluating any arbitrary visual encoder and representation in the context of general-purpose vision.


\section{G-VUE Benchmark}
\label{sec:gvue}
Mirroring human cognitive abilities, our benchmark is comprised of tasks from four functional domains that a general vision system should support: \textit{Perceive}, \textit{Ground}, \textit{Reason}, and \textit{Act}. \textit{Perceive} characterizes the basic ability of understanding geometry from raw visual input. \textit{Ground} connects geometry with concepts and semantics. \textit{Reason} operates on grounded objects and performs deductions. And finally, \textit{Act} makes decisions and plans actions out of extensive reasoning. For each domain, we select visual tasks and datasets that are \textbf{representative, challenging, from the real-world domain, and available to the public}. The full list of the tasks is shown in~\cref{tab:benchmark}. We provide an in-depth description in this section and \cref{sec:design} for task selection, dataset selection, and rationales. 

\def\arraystretch{1.5}
\begin{table}[t!]
    \centering
    \small
    \caption{The full list of tasks in G-VUE. We categorize our tasks according to their functional domains and provide details on the selected datasets, train/val/test splits, input/output modalities, and evaluation metrics.}
    \label{tab:benchmark}
    \resizebox{\linewidth}{!}{%
        \begin{tabular}{lcccccc}
        \toprule
        \textbf{Task}  & \textbf{Dataset} & \textbf{\#Train} / \textbf{\#Val} / \textbf{\#Test} & \textbf{Text} & \textbf{Output} & \textbf{Metrics} \\
        \midrule
        \multicolumn{6}{l}{\textbf{Perceive}} \\
        \hline
        Depth Estimation & NYUv2 & 24k / - / 0.6k & N & Depth Map & d\textless{}1.25, AbsRel, RMSE \\
        Camera Pose Estimation & CL \& 7-Scenes & (3.8k, 26k) / - / (1.1k, 17k) & N & Camera Pose & Mean Trans. \& Orient. Error\\
        3D Reconstruction & ShapeNetCore & 30k / - / 7.8k & N & Volumetric SDF & 3D IoU\\
        \hline
        \multicolumn{6}{l}{\textbf{Ground}} \\
        \hline
        Image-Text Retrieval & Flickr30k & 29k / 1.0k / 1.0k & Y & Matching Score & Recall@1,5,10 \\
        Phrase Grounding & RefCOCO & 42k / 3.8k / (2.0k, 1.8k) & Y & Bbox & Acc@0.5 \\
        Semantic Segmentation & ADE20k & 20k / 2.0k / - & N & Segmentation Map & mIoU \\
        \hline
        \multicolumn{6}{l}{\textbf{Reason}} \\
        \hline
        Question Answering & GQA & 943k / 132k / 12.5k & Y & Choice & Accuracy \\
        Commonsense Reasoning & VCR & 213k / 26.5k / - & Y & Choice & Accuracy \\
        Abstract Reasoning & Bongard-HOI & 23k / 17k / 14k & N & Binary Label & Accuracy \\
        \hline
        \multicolumn{6}{l}{\textbf{Act}} \\
        \hline
        Navigation & R2R & 14k / (1.0k, 2.3k) / 4.2k & Y & Next Move & SPL \\
        Manipulation & Ravens & 0.8k / 0.08k / 0.8k & Y & Pick \& Place & Success Score \\
        \bottomrule
        \end{tabular}
    }%
\vspace{-10px}
\end{table}


\subsection{Perceive}
Visual perception underlies human's subconscious mental approximation of geometric relationships in the physical world~\citep{clements2004geometry}. In the domain of \textit{Perceive}, we start with the evaluation of the sense of distance and space through \textbf{Depth estimation}. We move onto egocentric pose understanding with \textbf{Camera pose estimation}. Lastly, we examine the ability for detailed, shape-based geometric understanding with \textbf{Single-view 3D reconstruction}.


\paragraph{Depth estimation.} Monocular depth estimation is the task of estimating the egocentric distance to objects on a single 2D image. We adopt NYUv2 \citep{silberman2012indoor}, a common practice for evaluating monocular depth estimation. We make use of the updated version proposed by~\citet{lee2019big} as the training set to provide a sufficient amount of data samples. The model's performance is evaluated by relative and absolute errors.


\paragraph{Camera pose estimation.} Learning-based absolute camera pose estimation is an instance-level task that requires the model to infer parameters about the point of view from a scene's geometry.
Following the general setting in previous works \citep{kendall2015posenet}, we use both an outdoor urban localization dataset, Cambridge Landmarks \citep{kendall2015posenet}, and an indoor scenes dataset, 7-Scenes \citep{shotton2013scene} as our data source. We use the training and testing splits provided by \citet{kendall2015posenet}, and report the mean errors of position (meters) and orientation (degrees) respectively over the scenes as our metrics.

\paragraph{Single-view 3D reconstruction.} The task of single-view 3D reconstruction requires the model to predict a 3D voxel shape from a single 2D image. 
We evaluate this capability on the synthetic images \citep{choy20163d} of 3D objects from ShapeNetCore \citep{chang2015shapenet} and use the train/test split provided by \citet{xu2019disn}.
We use 3D IoU in $64 ^ 3$ resolution as the metric to evaluate the performance.

\subsection{Ground}
Visual grounding is defined as the process of connecting and aligning perceptual visual inputs with natural language. This alignment comes at different granularities, which form the basis of task progression in the \textit{Ground} domain. We start with image-level content matching in \textbf{Image-text retrieval}, and proceed onto instance-level content matching in \textbf{Phrase grounding}. We further consider the composition of both granularities, \ie, scene-level instance property grounding, in \textbf{Semantic segmentation}.

\paragraph{Image-text retrieval.} Image-text retrieval requires the model to match the image-text pairs given a sample from one modality. We select Flickr30k \citep{young2014image} for this task. We follow the dataset split in \citet{karpathy2015deep}, with 29k/1k/1k images for training, validation, and test sets respectively. We use the recall of finding the correct caption in top-1 (R@1), top-5 (R@5), and top-10 (R@10) predictions as the metric for retrieval.

\paragraph{Phrase grounding.} In phrase grounding, a model is asked to predict a bounding box that crops out an object referred to by a natural language phrase. We select RefCOCO \citep{yu2016modeling} as our source. It is a dataset built upon MSCOCO~\citep{lin2014microsoft} with objects and bounding boxes labeled for each image. We adopt the benchmarking splits of person reference (testA) and non-person reference (testB) in RefCOCO, in addition to the validation set.

\paragraph{Semantic segmentation.} Semantic segmentation requires the model to ground semantic labels to pixels in each image and make dense predictions. For this task, we select the curated version of ADE20K~\citep{zhou2017scene} and the corresponding train/test split from the MIT Scene Segmentation Benchmark~\citep{zhou2017scene}. It contains 150 categories of stuff (\eg, sky) and discrete objects (\eg, car) with their corresponding semantic masks on each image. We use the \ac{miou} between predicted and ground truth segmentation mask as the evaluation metric.

\subsection{Reason}

Visual reasoning characterizes the process of collecting grounded facts from images and making a complex deduction, \eg, answering questions. In \textit{Reason} tasks, we first visit \textbf{Visual question answering}, a seminal task that comprehensively examines whether a learner can jointly reason with images and text. We then advance to visual \textbf{Commonsense reasoning}, where a model also needs to acquire physical and societal commonsense through learning. Our final task examines a not often discussed niche in visual tasks: \textbf{Abstract and few-shot reasoning}.



\paragraph{Visual question answering.} In a VQA problem, the model needs to generate the answer to a question (a single word or a compound word) after viewing an image. We select GQA \citep{hudson2019gqa} for this task as it requires challenging relational and multi-step reasoning. By default, GQA provides image region features extracted with Faster-RCNN and ResNet-50. We instead use grid features output by the backbone directly without object detection. GQA also offers rich metrics for evaluation, but we only use the overall accuracy in our benchmark to ensure consistency with other tasks. Accuracy on the test-dev set is reported.

\paragraph{Commonsense reasoning.} This task aims at answering multiple-choice questions out of scenarios that require commonsense knowledge. To crack these puzzles, a learner needs to understand human states, activities, social norms, and other commonsense knowledge. We select VCR \citep{zellers2019recognition} dataset to evaluate commonsense reasoning with images.  In this dataset, we follow the data preprocessing pipeline in \citet{zellers2021merlot}, referring to entities by highlighting regions with pre-defined colors. For simplicity, we only incorporate the question answering (Q $\rightarrow$ A) subtask and report the overall accuracy on the validation set.


\paragraph{Abstract and few-shot reasoning.} We employ Bongard-HOI \citep{jiang2022bongardhoi} as a proxy to evaluate the ability of abstract and few-shot reasoning. The task can be briefly described as follows: given a small set of positive and negative images that either depict or don't depict a human-object interaction (HOI) concept, the objective is to determine whether the additionally provided query image depicts that HOI concept. Bongard-HOI offers four val and test sets, respectively, which comprise seen/unseen objects and seen/unseen actions, allowing systematic evaluation of the model's generalization capability. We follow the evaluation protocol introduced by \citet{jiang2022bongardhoi} and evaluate the overall accuracy on the four test sets.


\subsection{Act}

\textit{Act} is the most complex function vision systems could support. We consider two aspects of acting: navigating in an environment and manipulating objects. Both tasks require extensive reasoning about the environment and planning accordingly. Specifically, \textbf{Navigation} proposes the challenge of identifying one's location, understanding their possible movements, and directing to the target location from an egocentric perspective. \textbf{Manipulation}, on the other hand, focuses on learning from allocentric-view observations for planning with granular visual-motor skills.




\paragraph{Navigation.} We adopt the Matterport3D Simulator (MatterSim) \citep{mattersim} and its associated Room to Room (R2R) dataset as the setting for vision-language navigation (VLN). This task requires the agent to follow natural language instructions and traverse a virtual house on a mesh formed by viewpoint locations, each with a panoramic view of the surroundings. Following previous works, we evaluate the performance on validation subsets and use SPL \citep{anderson2018spl} in the unseen environment as the primary metric. 

\paragraph{Manipulation.} We follow CLIPort \citep{shridhar2022cliport} and adopt their environment for the manipulation task. This task requires the agent to perform compositional actions (\eg, pick and place) based on visual observations and language instructions. As in \citet{shridhar2022cliport}, visual modules make affordance predictions to guide the subsequent action module on where and how to manipulate. We collect 8 subtasks with seen/unseen splits out of the 10 initial language-goal-conditioned subtasks proposed by \citet{shridhar2022cliport} and perform affordance learning on seen split of subtasks jointly, with 100 demonstrations for each task. We adopt the success scores used in \citet{zeng2020transporter} and report average scores on the unseen subsets of subtasks for evaluation.

\section{Framework}\label{sec:model}

Due to the diverse nature of tasks in \ac{bm}, evaluating visual representations could be challenging. To this end, we propose an encoder-decoder framework for the visual tasks, making it possible to adapt arbitrary visual representations to all 11 tasks.

An overview of our framework can be found in \cref{fig:framework}. Firstly, the visual representation is viewed as the output of an image encoder, \eg, ResNet-50. For tasks that also incorporate textual input, a text encoder is used to produce appropriate language embedding. Specifically, we employ RoBERTa \citep{liu2019roberta} for language-involved tasks. Then the visual representation (and possibly the language embedding) will be sent to a task decoder. Since the tasks in \ac{bm} require various formats of output, we design different decoders for each class of output format. We briefly summarize the variants of our decoder as follows. More details are deferred to \cref{sec:implementation}.

\begin{itemize}[leftmargin=*,noitemsep,nolistsep]
    \item \textbf{Label decoder.} This decoder is composed of several Transformer blocks and consumes tokenized visual representation (possibly concatenated with text embedding). We insert an extra \texttt{[CLS]} token for prediction. It works with most of the classification and regression tasks in \ac{bm}.
    \item \textbf{Dense decoder.} This decoder adopts the Segformer head \citep{xie2021segformer}, a lightweight module that maps tokenized visual representation to dense output. We use it in dense prediction tasks.
    \item \textbf{3D decoder.} This decoder resembles the pipeline in \citet{mittal2022autosdf}, where a Transformer-based decoder maps tokenized visual representation to a joint distribution over shapes and finally produces volumetric SDF via a pre-trained VQ-VAE. It is only invoked in the 3D reconstruction task.
    \item \textbf{Navigation decoder.} This decoder follows \citet{hong2021vln}, which maps visual representations and instruction embeddings to movement actions in a recurrent manner.
    \item \textbf{Manipulation decoder.} We employ the design in \citet{shridhar2022cliport} and map the visual representation and instruction embeddings to an affordance map, which then guides a model-based planner to draw the next manipulation action.
\end{itemize}

Albeit our efforts to make the encoder-decoder framework more generic and powerful, we do acknowledge that many task-specific decoders could generate better results given the same visual representation. However, we would like to highlight that our design principle is to balance performance and the amount of inductive bias, as the purpose of \ac{bm} is to provide a generic and accessible platform for evaluating general-purpose visual representation. Architectures with too many complex and task-specific designs are not consistent with the design principle and long-term goal of \ac{bm}. More discussions can be found in the Appendix.

\section{Experiment}

\subsection{Visual Representations for Evaluation}
We analyze three factors that may affect the quality of visual representation: architecture, pre-training mechanism, and source data. In our experiments, we adopt two architectures commonly used in large-scale pretraining: ResNet \citep{he2016deep} and ViT \citep{dosovitskiy2020image}. On top of these backbones, we consider variations of pre-training mechanisms including the extent of supervision (\eg, supervised \vs self-supervised) and learning objectives (\eg, discriminative \citep{he2020momentum} \vs generative \citep{he2021masked}). For source data, we consider ImageNet \citep{deng2009imagenet}, large-scale image-caption pairs \citep{radford2021learning}, and Ego4D videos \citep{grauman2021ego4d}. To fully explore the effect of these factors, we select 7 representative visual representations as shown in~\cref{tab:backbone}.
\begin{table}[t!]
    \centering
    \caption{An overview of evaluated visual representations.}
    \label{tab:backbone}
    \resizebox{0.8\linewidth}{!}{%
        \begin{tabular}{cccc}
        \toprule
        Representation & Architecture & Pre-training mechanism & Data \\
        \midrule
        RN-IN & ResNet-50 & Supervised classification & ImageNet \\
        RN-MoCo & ResNet-50 & Self-supervised Contrastive Learning & ImageNet \\
        RN-CLIP & ResNet-50 & Vision-language Contrastive Learning & WebImageText \\
        RN-Ego & ResNet-50 & Vision-language Contrastive Learning & Ego4D \\
        ViT-32-CLIP & ViT-B/32 & Vision-language Contrastive Learning & WebImageText \\
        ViT-16-CLIP & ViT-B/16 & Vision-language Contrastive Learning & WebImageText \\
        ViT-16-MAE & ViT-B/16 & Self-supervised Masked Image Modeling & ImageNet \\
        \bottomrule
        \end{tabular}
    }%
\end{table}

{
\newcommand{\sepline}{\hline}
\renewcommand{\arraystretch}{1.2}
\begin{table}[t!]
\caption{Quantitative results of visual representations on G-VUE. For space considerations, we use the abbreviation of words for identifying tasks (\eg ``Cam. Pose.'' for camera pose estimation, ``I-T Retr.'' for image-to-text retrieval, ``Phr. Grnd.'' for phrase grounding, ``Sem. Seg.'' for semantic segmentation, ``Com. Res'' for common sense reasoning, ``Abs. Res.'' for abstract reasoning, ``Nav.'' for navigation and ``Manip.'' for manipulation). Summary scores are attached at the bottom, which only consider test-dev subset for score on VQA task, test subset for score on abstract reasoning task, and unseen SPL for score on navigation task. We highlight the best results on each task both in bold and with underline.}
\label{tab:all_results}
\vskip 0.2in

\resizebox{\linewidth}{!}{
\begin{tabular}{c|c|ccccccc}
\toprule
\multicolumn{2}{c}{\multirow{2}[2]{*}{\textbf{Task}}} & \multicolumn{7}{c}{Representation} \\
\cmidrule(lr){3-9}
\multicolumn{2}{c}{} & RN-IN & RN-MoCo & RN-CLIP & RN-Ego & ViT-32-CLIP & ViT-16-CLIP & ViT-16-MAE \\
\hline
\multicolumn{2}{c}{\textbf{Perceive}} &  &  &  &  &  &  &  \\
\hline
\multirow{3}{*}{Depth Est.} & test d\textless{}1.25 $\uparrow$ & 0.6394 & 0.6400 & 0.6523 & 0.4548 & 0.7363 & 0.7608 & \underline{\textbf{0.7699}} \\
 & test AbsRel $\downarrow$ & 0.2190 & 0.2184 & 0.2097 & 0.3409 & 0.1699 & 0.1606 & \underline{\textbf{0.1554}}\\
 & test RMSE $\downarrow$ & 0.7292 & 0.7284 & 0.6987 & 1.0302 & 0.5888 & 0.5535 & \underline{\textbf{0.5486}} \\
 \cline{2-9}
 \multirow{4}{*}{Cam. Pose} & Trans.(CL) $\downarrow$& 2.909 & \underline{\textbf{1.972}} & 2.446 & 3.045 & 2.160 & 2.196 & 2.177 \\
 & Orient.(CL) $\downarrow$& 7.298 & 5.427 & 6.814 & 7.000 & 5.974 & 6.020 & \underline{\textbf{4.911}} \\
 & Trans.(7S) $\downarrow$& 0.281 & \underline{\textbf{0.241}} & 0.279 & 0.266 & 0.290 & 0.263 & 0.271 \\
 & Orient.(7S) $\downarrow$& 10.299 & 8.797 & 9.502 & 9.058 & 9.682 & 9.752 & \underline{\textbf{6.699}} \\
 \cline{2-9}
 3D Recon. & test mIoU $\uparrow$& 39.70 & 41.98 & 42.00 & 37.66 & 40.63 & 39.43 & \underline{\textbf{43.95}} \\
 \sepline
\multicolumn{2}{c}{\textbf{Ground}} &  &  &  &  &  &  &   \\
\hline
 \multirow{3}{*}{I-T Retr.} & test R@1 $\uparrow$& 25.2 & 23.3 & 42.4 & 2.6 & 47.0 & \underline{\textbf{59.5}} & 23.4 \\
 & test R@5 $\uparrow$& 51.9 & 48.8 & 71.2 & 10.6 & 76.1 & \underline{\textbf{86.0}} & 50.5 \\
 & test R@10 $\uparrow$& 64.5 & 60.0 & 82.4 & 16.9 & 85.3 & \underline{\textbf{92.5}} & 63.6  \\
  \cline{2-9}
 \multirow{3}{*}{Phr. Grnd.} & val Acc@0.5 $\uparrow$& 48.57 & 54.40 & 63.45 & 36.5 & \underline{\textbf{67.99}} & 64.55 & 65.18 \\
 & testA Acc@0.5 $\uparrow$& 54.48 & 59.14 & 70.63 & 41.92 & \underline{\textbf{75.70}} & 70.73 & 67.59 \\
 & testB Acc@0.5 $\uparrow$& 43.87 & 50.88 & 55.86 & 31.38 & 61.16 & 57.24 & \underline{\textbf{62.10}} \\
 \cline{2-9}
 Sem. Seg. & val mIoU $\uparrow$& 18.83 & 18.05 & 22.32 & 5.00 & 29.14 & \underline{\textbf{35.86}} & 26.25 \\
\sepline
\multicolumn{2}{c}{\textbf{Reason}} & & & & & & & \\
\hline
\multirow{2}{*}{VQA} & val Acc. $\uparrow$ & 52.79 & 50.20 & 56.09 & 45.56 & \underline{\textbf{59.41}} & 59.18 & 54.82 \\
 & test-dev Acc. $\uparrow$ & 47.54 & 45.02 & 50.01 & 41.20 & \underline{\textbf{51.88}} & 51.65 & 48.50 \\
 \cline{2-9}
 Com. Res. & val Acc. $\uparrow$ & 52.61 & 50.98 & 57.30 & 52.79 & 61.94 & \underline{\textbf{64.11}} & 60.76 \\
 \cline{2-9}
 \multirow{2}{*}{Abs. Res.} & val Acc. $\uparrow$ & 59.85 & 62.38 & 64.54 & 55.72 & 63.24 & \underline{\textbf{65.65}} & 62.12 \\
 & test Acc. $\uparrow$ & 62.14 & 64.63 & 67.32 & 56.09 & 65.55 & \underline{\textbf{69.43}} & 62.85 \\
\sepline
\multicolumn{2}{c}{\textbf{Act}} &  &  &  &  &  &  &   \\
\hline
\multirow{2}{*}{Nav.}
& Unseen Succ. $\uparrow$& 48.62 & 48.49 & 49.00 & 43.89 & 49.21 & 47.13 & \underline{\textbf{49.55}}\\
& Unseen SPL $\uparrow$& 42.61 & 43.49 & 43.49 & 37.88 & \underline{\textbf{43.99}} & 43.60 & 43.72 \\
\cline{2-9}
Manip. & Unseen Score $\uparrow$& 32.19 & 36.94 & 37.47 & 27.78 & 28.80 & \underline{\textbf{40.80}} & 37.98 \\
\hline
\multicolumn{2}{c}{\textbf{Summary Score}} & 42.05 & 43.15 & 47.18 & 33.29 & 48.31 & \underline{\textbf{51.39}} & 46.85 \\
\bottomrule
\end{tabular}
}
\end{table}
}

During training, we keep these representations fixed and tune task-specific decoders for each task. Notably, we mainly chose grid features in our evaluated visual representations, as it is more generic, flexible for adaptation, and efficient compared with object-centric features~\citep{jiang2020defense,kim2021vilt}. Recent studies~\citep{carion2020end,caron2021emerging,ma2022relvit} further shows emerging object-centric features in grid feature learning, which reveals their potential in tasks (\eg VQA) where object-centric features are commonly adopted.




\subsection{Results and Analyses}

We present quantitative experiment results in \cref{tab:all_results}. In addition to metrics on each task, we provide a \textit{summary score} (the bottom line of \cref{tab:all_results}) as an overall measure across all tasks by converting all error metrics into percentages and taking an average. We summarize our major findings as follows.


\paragraph{Architecture.} In general, \textit{\textbf{ViT outperforms ResNet on most visual tasks.}} This confirms similar observations made by a few recent works~\citep{liu2021swin,wang2021pyramid,shen2021much,ma2022relvit}. 
Specifically, we find ViT-32-CLIP and ViT-16-CLIP significantly outperform RN-CLIP on dense prediction tasks, \ie, depth estimation and semantic segmentation. This may indicate Transformer architecture is endowed with better compatibility on fine-grained discriminative tasks. 

\paragraph{Pre-training mechanism.}
We compare three pre-training mechanisms: supervised learning, self-supervised learning, and vision-language contrastive learning. Supervised learning on ImageNet was once the most common practice for producing general visual representations, but it is recently challenged by the simple yet effective self-supervised learning~\citep{he2020momentum,chen2020simple}. We observe similar results on \ac{bm}. Representation from supervised pre-training (RN-IN) is only better than self-supervised one (RN-MoCo) in a handful of tasks that require semantics (\eg, image-text retrieval) while significantly inferior to self-supervised one in tasks highly dependent on geometric perception (\eg, camera pose estimation). 

Vision-language contrastive learning is another competitive pre-training scheme equipped with semantics alignment. Due to the contribution of language, visual representation obtained in this way shows considerably stronger generalizability regardless of the lack of direct supervision~\citep{radford2021learning}. This is demonstrated by the observation that \textit{\textbf{CLIP visual representation dominates most V-L tasks, and achieves the highest summary score.}} 


\paragraph{Source training data.} The experimental results also indicate that, on visual tasks that require semantics, especially language alignment, the gap between representations trained from supervised learning and self-supervised learning on ImageNet could be less significant than the gap with CLIP visual representations. We hypothesize that the cause lies in the source training data. It appears that ImageNet is already insufficient to support the learning of a general-purpose visual representation that can accommodate the broad spectrum of tasks in \ac{bm}. On the other hand, \textit{\textbf{the success of CLIP visual representations in \ac{bm} supports the recent paradigm shift to multi-modal pre-training,}} which leverages massive source data (\eg, WebImageText) and facilitates better representation with grounded concept learning. Such large-scale pre-training provides a promising path towards general-purpose visual representation and could potentially help to solve \ac{bm}. As a side note, we also notice that visual representations learned from egocentric video clips (RN-Ego) fail to achieve comparable performances on \ac{bm}. The drastic domain shift from the hand-centered Ego4D data to more general natural images in \ac{bm} might be the primary cause for such a performance drop.


\paragraph{Language representation.} To make the evaluation of visual representations fair, we provide them with an identical language representation, \ie, RoBERTa as language encoder. 
Nevertheless, using language representation pre-trained on the text-only corpus to collaborate with visual representations may cause sub-optimal performances, see~\cref{sec:ablation}. This is similar to the shortcoming of vision-only pre-training. Check \cref{sec:ablation} for detailed ablation studies on the language representation.


In addition, we explore some other factors that could improve the performances of visual representations on \ac{bm}, including fine-tuning and resolution. We summarize these studies in \cref{sec:ablation}.


\subsection{Task Correlation and Generalization}
\begin{wrapfigure}{r}{0.485\linewidth}
  \begin{center}
    \includegraphics[width=\linewidth]{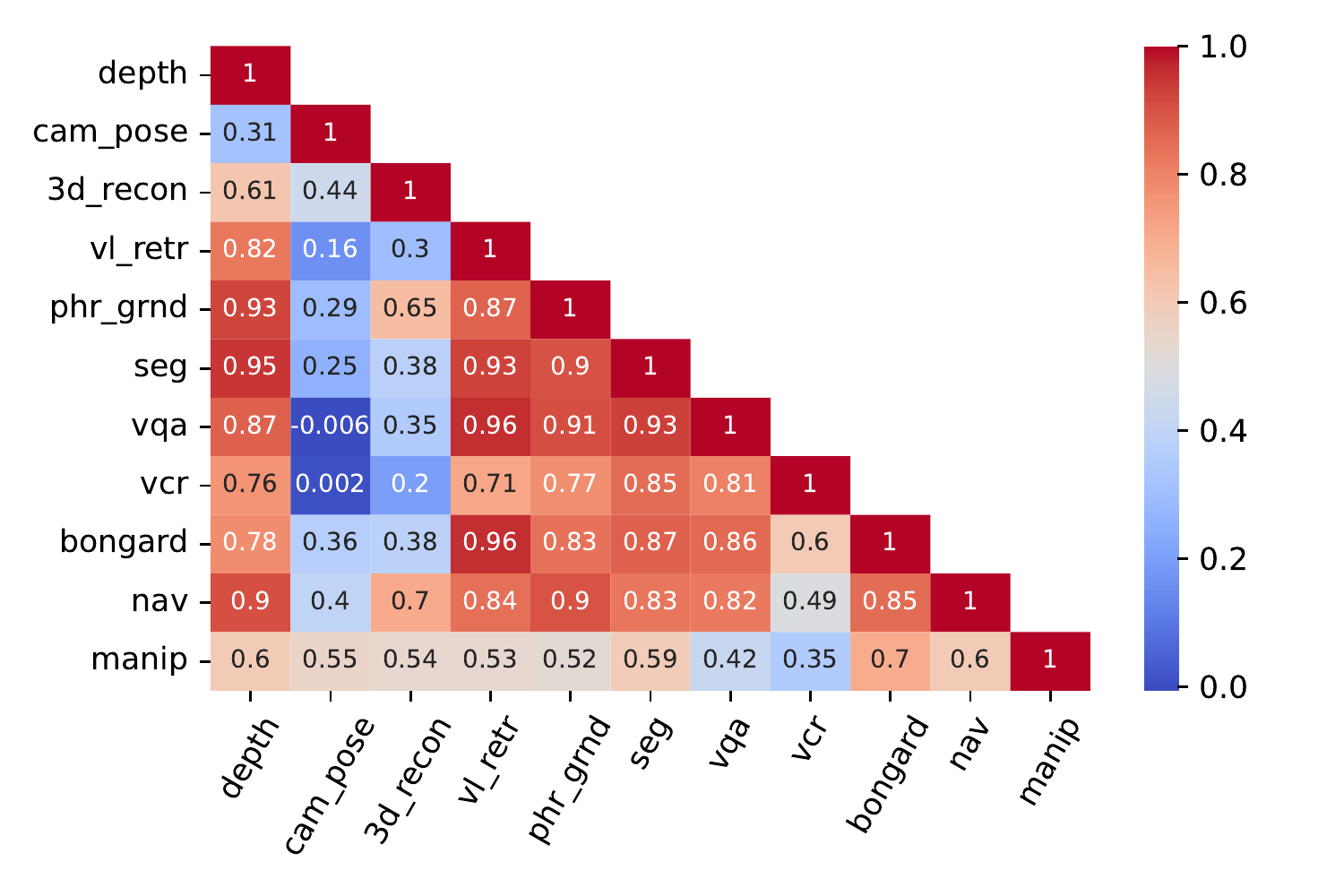}
  \end{center}
  \vspace{-10px}
  \caption{Correlations of 11 tasks in G-VUE.}
  \label{fig:correlation}
\end{wrapfigure}
It is believed difficult for visual representations learned from specific tasks to adapt to other domains. 
However, as our results suggest, this is not necessarily the case (\eg MAE models perform surprisingly well on \textit{Ground} and \textit{Reason} tasks with reconstruction objectives).
Such findings further motivate the question: are there shared requirements on representations over the seemingly divergent tasks? 
As there is no clear relationship between tasks, we empirically estimate task correlations with our experimental results. Specifically, we construct a task correlation matrix by treating the representation's performance on each task as a variable. 
As shown in \cref{fig:correlation}, we observe that most tasks show high correlations. In contrast to other tasks, camera pose estimation and 3D reconstruction show fewer connections to other tasks. Nonetheless, camera pose estimation still correlates with manipulation since they both require 3D geometric understanding. We also observe significant correlations in \textit{Ground} and \textit{Reason} domain. This result is intuitive since these tasks share similar objectives. We hope these findings could foster new insights into multi-task learning.

On the other hand, we find that visual representations may not be able to generalize across large modality gaps. RN-Ego is trained with first-person videos that have drastically different viewpoints compared to other data sources. As a result, we observe inferior performance of RN-Ego on \ac{bm} compared with other image-based representations. Additionally, we recognize that visual representations learned with strong biases are not suitable for generalizing across tasks, \eg, representation fine-tuned on depth prediction shows significant performance drops when transferring to other tasks despite the high task correlation.

\subsection{Evaluation of Latest Models}
Recently there emerge more advanced general-purpose models~\citep{li2022grounded,wang2022unifying,lu2022unified} from large-scale pre-training. To provide a more complete view on these models, we evaluate them on G-VUE and answer two questions: 1) how different are visual representations with advanced pre-training, 2) how will unified V-L models perform under holistic evaluation. Specifically, we select GLIP-tiny~\citep{li2022grounded} that uses Swin Transformer tiny~\citep{liu2021swin} for visual representation. We select the largest model variant of OFA~\citep{wang2022unifying} and UIO~\citep{lu2022unified} respectively and directly make inference on tasks they can solve, since they are highly integrated without explicit visual representation. \cref{tab:add_eval} shows the results.

\begin{table}[t!]
    \centering
    \caption{Evaluation of latest general-purpose models on G-VUE. For GLIP-tiny, we take the pre-trained visual representation and adapt it for various tasks like the aforementioned baselines. ``*'' means OFA and UIO directly perform inference on G-VUE. ``$\dagger$'' indicates the performance of UIO on VQA lags behind probably because it is not exposed to data domain like GQA. ``$\ddagger$'' explains the reason to account for sub-optimal performance on VCR is that OFA and UIO have no chance to adapt to the setting of character-region reference.}
    \vskip 0.1in
    \label{tab:add_eval}
    \resizebox{1.0\linewidth}{!}{%
        \begin{tabular}{cccccccccccc}
        \toprule
        \multirow{2}{*}{Model} & \multicolumn{3}{c}{Depth} & \multicolumn{2}{c}{Cam. Pose} & 3D Recon. & \multicolumn{3}{c}{I-T Retr.} \\
        \cmidrule(lr){2-4}\cmidrule(lr){5-6}\cmidrule(lr){7-7}\cmidrule(lr){8-10}
         & d<1.25 & AbsRel $\downarrow$ & RMSE $\downarrow$ & Trans. / Orient. (CL) & Trans. / Orient. (7S) & mIoU & R@1 & R@5 & R@10 \\
        \midrule
        GLIP-Tiny & 0.7195 & 0.1807 & 0.6116 & 1.962 / 8.099 & 0.326 / 11.764 & 41.57 & 26.3 & 56.2 & 68.3 \\
        OFA-Huge$^*$ & - & - & - & - & - & - & 26.1 & 46.3 & 54.9 \\
        UIO-XL$^*$ & 0.9439 & 0.0860 & 0.3505 & - & - & - & 33.0 & 53.9 & 62.8 \\
        \bottomrule
        \end{tabular}
    }%
    \vspace{7pt}
    \resizebox{1.0\linewidth}{!}{%
        \begin{tabular}{cccccccccccc}
        \toprule
        \multirow{2}{*}{Model} & \multicolumn{3}{c}{Phr. Grnd. / Acc@0.5} & Sem. Seg. & VQA & Com. Res. & Abs. Res. & \multicolumn{2}{c}{Nav. (Unseen)} & Manip. (Unseen) \\
        \cmidrule(lr){2-4}\cmidrule(lr){5-5}\cmidrule(lr){6-6}\cmidrule(lr){7-7}\cmidrule(lr){8-8}\cmidrule(lr){9-10}\cmidrule(lr){11-11}
         & val & testA & testB & mIoU & Acc. & Acc. & Acc. & Succ. & SPL & Score \\
        \midrule
        GLIP-Tiny & 65.31 & 69.27 & 59.94 & 28.19 & 47.39 & 61.10 & 65.73 & 47.98 & 43.73 & 44.37 \\
        OFA-Huge$^*$ & 92.04 & 94.03 & 88.44 & - & 59.92 & 28.19$^\ddagger$ & - & - & - & - \\
        UIO-XL$^*$ & 86.17 & 86.28 & 85.30 & - & 36.91$^\dagger$ & 32.09$^\ddagger$ & - & - & - & - \\
        \bottomrule
        \end{tabular}
    }%
\vspace{-12pt}
\end{table}



Regarding the first question, the results show that grounded V-L pre-training~\citep{li2022grounded} can boost the lightweight visual backbone to reach comparable performances with CLIP baselines, and even outperforms on manipulation task by a large margin. This indicates visual representation from object-centric pre-training can benefit embodied perception and manipulation as well.

On the other hand, the results demonstrate the remarkable performance of recent unified models, mainly on tasks where they have been pre-trained. For example, UIO achieves superior performance on depth estimation, and both of OFA and UIO reach the state-of-the-art on phrase grounding. However, they are still in lack of versatility to deal with diverse tasks in G-VUE, \eg, 3D reconstruction, navigation and manipulation. These visual tasks are substantial and form the core of human visual system. Hence, we advocate the general vision model should expand to a broader domain.

\section{Conclusion}

In this work, we present a novel general-purpose vision benchmark \ac{bm}, consisting of 11 meticulously chosen tasks. \ac{bm} covers the full spectrum of visual skills over four domains: \textit{Perceive}, \textit{Ground}, \textit{Reason} and \textit{Act}. We further introduce an encoder-decoder framework that supports the evaluation of arbitrary visual representation on all 11 tasks. With \ac{bm}, we evaluate the performance of representative visual representations pre-trained with different learning paradigms and data sources. In particular, we find that Transformer-based architectures beat CNN-based models in most visual tasks and that pre-training on massive image-text pairs such as CLIP achieves significantly better results than ImageNet pre-training. Such findings shed light on the path to building a unified model for general-purpose vision. 

We will also open-source an evaluation platform with a public leaderboard that can fairly and holistically evaluate different models on G-VUE. We hope this effort will promote the study of general-purpose representation learning and encourage the computer vision community to pursue universal, easily adaptable, and general-purpose visual models.

\bibliography{ref}
\bibliographystyle{unsrtnat}

\appendix
\clearpage

\section{Benchmark Overview}
\begin{figure}[h!]
\centering
 \includegraphics[width=\linewidth]{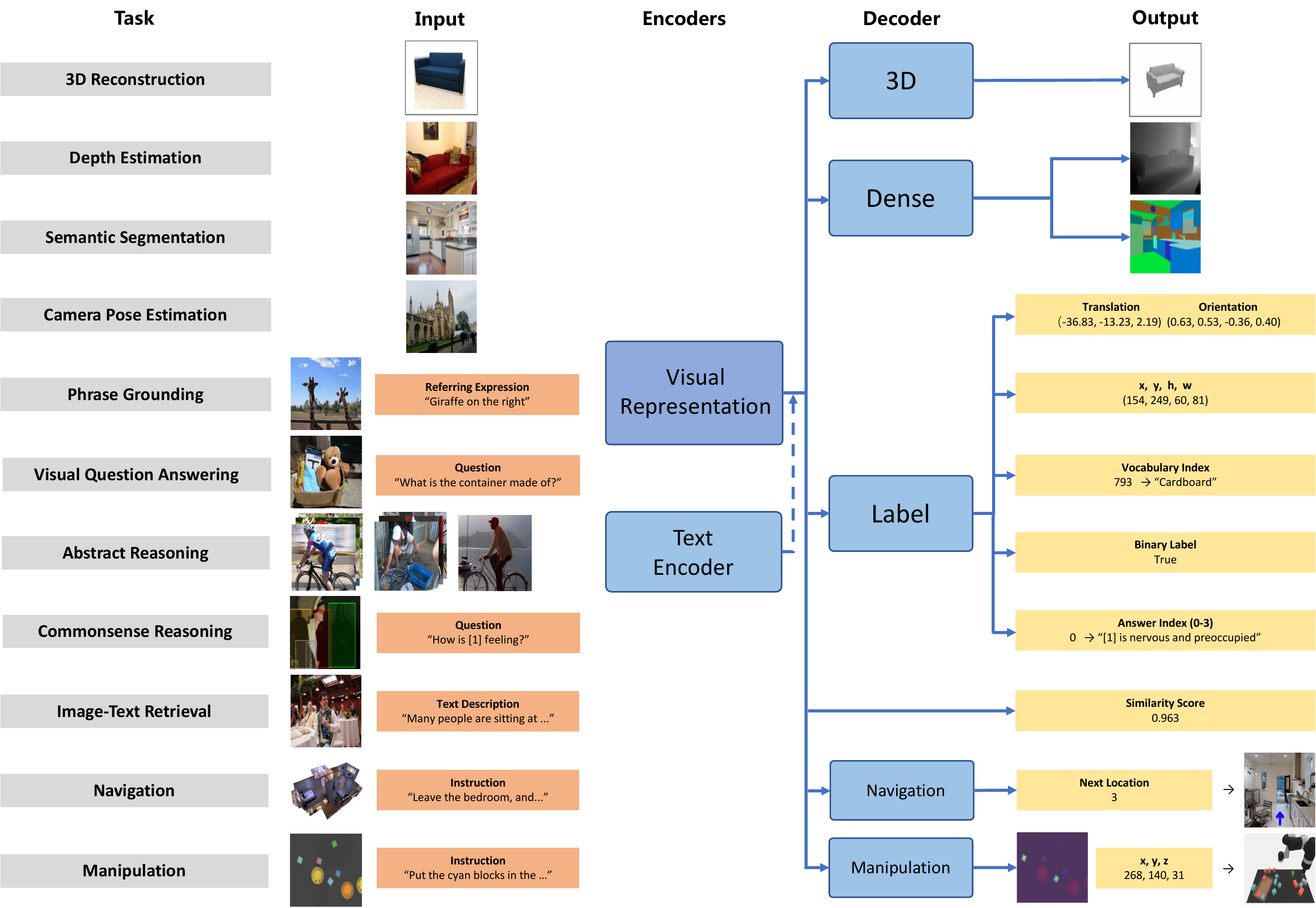}
\caption{An illustration of our encoder-decoder framework. The visual representation is the output of an image encoder, e.g. ResNet-50. A text encoder is used to produce an embedding of the textual input for tasks that also incorporate a textual input. The visual representation, possibly with the text embedding, will be sent to task decoders to accomplish the corresponding tasks.}
\label{fig:framework}
\end{figure}

To present an illustration of our benchmark, we gather the inputs and outputs of all 11 tasks and exhibit how the pipeline works in a holistic layout. See \cref{fig:framework}.

\paragraph{Online evaluation and leaderboard.} We host an online evaluation platform and a leaderboard for the G-VUE benchmark on EvalAI \citep{yadav2019evalai}. More details can be found through our project website at \href{https://sites.google.com/view/g-vue/}{https://sites.google.com/view/g-vue/}.

\paragraph{Code release.} We provide a suite of dataloaders, models, and scripts for training and evaluating visual representations on G-VUE. The repository is released at \href{https://github.com/wllmzhu/G-VUE}{https://github.com/wllmzhu/G-VUE}.

\section{Design Principles and Rationality} \label{sec:design}
To present the details of selecting tasks and datasets, we separate the four domains and describe each domain by first delivering the motivations behind task selections and then providing rationales for dataset choices.

\subsection{Perceive}
\paragraph{Task.} Depth perception is a foundational basis for projecting the 2D observation to 3D, which makes it essential in geometric understanding. As such, this constitutes the first task in the \textit{Perceive} domain. Specifically, we select monocular depth estimation as our examined task. Tasks like surface normal estimation regarding geometric understanding are alternatives to our selection. Due to their similar formulations, we select depth estimation as a representative and refer readers to \citet{zamir2018taskonomy} for a more in-depth summary of mid-level visual representations. In addition to depth estimation, we augment the \textit{Perceive} domain with the camera pose estimation task as it tests models' capabilities in estimating the pose of an agent. Camera pose estimation is also crucial for downstream applications, \eg, egocentric vision and robotics.

With depth and camera pose, we are able to recover the global geometric cues from a macroscopic perspective. Next, we examine detailed object-centric perception. One task candidate is 6D pose estimation, which requires the model to estimate both the location and the orientation of objects. However, it ignores some key components in perception, \eg, the appearance and shape of objects. We argue that geometric perception should consider these features, and to cover them we decide to choose the single-view 3D object reconstruction task. Furthermore, the capability of reconstructing objects can contribute to 3D holistic scene reconstruction. Therefore, this task can be regarded as a comprehensive test on \textit{Perceive} domain.

\paragraph{Dataset.} We choose the NYUv2 dataset for evaluation on the depth estimation task. Compared with another commonly used dataset KITTI \citep{geiger2012we} for autonomous driving, NYUv2 covers more curated indoor scenes, which is also consistent with tasks in \textit{Act} domain. We choose the 7-Scenes dataset and the Cambridge Landmarks dataset for evaluation on camera pose estimation. The former contains seven small-scale indoor scenes with only a few square meters and the latter includes six large-scale outdoor scenes whose space ranges from 35x25m (Shop Façade) to 500x100m (Street). This choice comes from two aspects: (1) they are widely adopted by previous works, (2) they encompass typical scenarios and challenges associated with estimating camera pose, \eg, scales, indoor scenes \vs outdoor scenes, and varying viewpoints. We select the ShapeNetCore dataset for evaluation on the 3D reconstruction task since it provides sufficient rendering tools and is widely adopted for single-view 3D reconstruction.

\subsection{Ground}
\paragraph{Task.} Current vision-language tasks involve semantics grounding on various granularities, \eg, image-level and pixel-level. This indicates a hierarchy and inspires us to span the \textit{Ground} domain via incremental granularities of vision-language grounding. The coarsest granularity lies on the image-sentence level. Image-sentence pairs are also the most accessible vision-language data, and identifying these pairs reflects the basic capability of grounding. Naturally, we select the image-text retrieval task to evaluate this capability.

Though, the capability to match image-text pairs cannot guarantee correct spatial grounding. An ideal vision model needs to tell not only ``what'', but also ``where''. Therefore, we shall consider the evaluation of object localization. To evaluate this capability, we select the phrase grounding task rather than a conventional detection task, since we care more about precise recognition aligned with language. With the phrase grounding task, we attain the granularity on the instance-level in \textit{Ground} domain. Given the phrase-to-object setting here, to avoid repeated text-to-image manner, we adopt the image-to-text setting in the aforementioned image-text retrieval task.

The phrase grounding task provides detailed alignment on the object-phrase level but makes the rectangle boundary assumption for objects, which limits the flexibility of grounded shape. Hence, we seek a task aimed at tagging each pixel in order to introduce a more flexible grounding. We select semantic segmentation (scene-level) rather than referring expression segmentation (object-level) since we hope to evaluate grounding in a more holistic fashion. To this end, with the three selected tasks, we enrich diversities in \textit{Ground} with various granularities and skills. Image captioning and human-object interaction (HOI) are also in our consideration. However, the former is similar to image-text retrieval and the latter can be found in Bongard-HOI.

\paragraph{Dataset.} We first determine the choice of RefCOCO for the phrase grounding task. RefCOCO, consisting of scenarios where one needs to discriminate between similar instances according to subtle hints in language descriptions, was proposed particularly for phrase grounding. RefCOCO+~\citep{yu2016modeling} and RefCOCOg~\citep{yu2016modeling} are closely related to RefCOCO, and picking one of them is enough. Next, for the image-text retrieval task, We choose Flickr30k rather than MSCOCO~\citep{lin2014microsoft} in order to reduce data overlap with RefCOCO. We choose ADE20K for semantic segmentation because it is more challenging compared with other counterparts \eg, PASCAL~\citep{everingham2010pascal}, Cityscapes~\citep{cordts2016cityscapes}. ADE20K also includes rich content such as stuff annotations.

\subsection{Reason}
\paragraph{Task.} This domain is designed to evaluate visual models for their capabilities of reasoning. As the evaluation of reasoning capability commonly comes in the form of question answering, we consider visual tasks that require answering questions regarding various scopes, starting from basic object abstraction and relation inference, to commonsense reasoning, and finally the abstract analogical reasoning between sets of images. Specifically, we first select a task evaluating the general capability of reasoning (\eg relations, counting), which is often referred to as visual question answering. Next, we select the commonsense reasoning task that focuses on the evaluation of cognitive capability, especially social knowledge. Furthermore, to introduce reasoning evaluation on a more challenging level, we choose the Bongard problem, which integrates abstract and analogical reasoning in a few-shot regime.


\paragraph{Dataset.}
To address the aforementioned critical dimensions of visual reasoning in the real-image domain, we select GQA, VCR, and Bongard-HOI as our benchmarking datasets. Specifically, (1) we prefer the GQA dataset over another commonly adopted dataset VQAv2 since GQA incorporates more complex compositional reasoning over attributes and relationships of objects; (2) VCR dataset focuses on commonsense knowledge; (3) Bongard-HOI was proposed for solving the Bongard problem in the context of human-object interaction (HOI), where one needs to abstract a key concept of action from two sets of images (few-shot) and infer whether this concept holds in query images.

\subsection{Act}
\paragraph{Task.} For this domain, we consider visual tasks in an embodied setting. The navigation task asks an agent to first localize and understand its positioning in an environment using visual information, and then plan its movements to the desired place. The manipulation task, on the other hand, requires the agent to achieve the desired configuration of the environment by planning the movement of environmental elements. We assume these two tasks sufficiently cover various skills required by embodied agents and therefore choose them as representative tasks in \textit{Act} domain.

\paragraph{Dataset.} For navigation, we adopt the Matterport3D dataset and simulator to train and evaluate virtual agents. We choose this dataset because it covers a real-world image domain with high visual complexity and diverse clean indoor environments. Matterport3D is also frequently used for vision-language navigation compared to other navigation datasets like Habitat~\citep{savva2019habitat}. For the manipulation task, we utilize the dataset of CLIPort that extends the Ravens benchmark~\citep{zeng2020transporter}. This dataset is equipped with rich variations of objects to facilitate the systematic evaluation of robotic manipulation. In addition, Matterport3D and CLIPort both provide seen and unseen splits for generalization test.

\section{Implementation Details}\label{sec:implementation}
\paragraph{Visual backbones.} We use Timm \citep{rw2019timm} to build most of the visual backbones. Feature extractors are manually implemented if not supported. To obtain CNN-like feature pyramids from ViT backbones, we simply apply bilinear interpolation on intermediate layers as it shows little difference compared with transposed convolution. All ResNets are ResNet-50, and all ViTs are ViT-Base.

\paragraph{Decoders.} We design three types of decoders for tasks other than \textit{Act}, and their architectures are visualized in \cref{fig:decoder}. Tasks in \textit{Act}, \ie navigation and manipulation, are performed by integrated agents in simulation environments, whose structures can be found in \citet{hong2021vln} and \citet{shridhar2022cliport}. The details of the decoders are as follows:
\begin{figure}[ht!]
    \centering
     \begin{subfigure}[b]{0.45\linewidth}
         \centering
         \includegraphics[width=\linewidth]{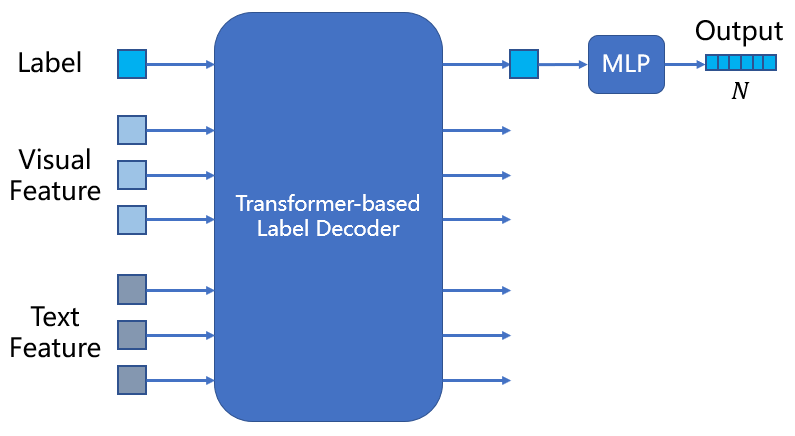}
         \caption{Label decoder.}
         \label{fig:label_decoder}
     \end{subfigure}
     \hfill
     \begin{subfigure}[b]{0.5\linewidth}
         \centering
         \includegraphics[width=\linewidth]{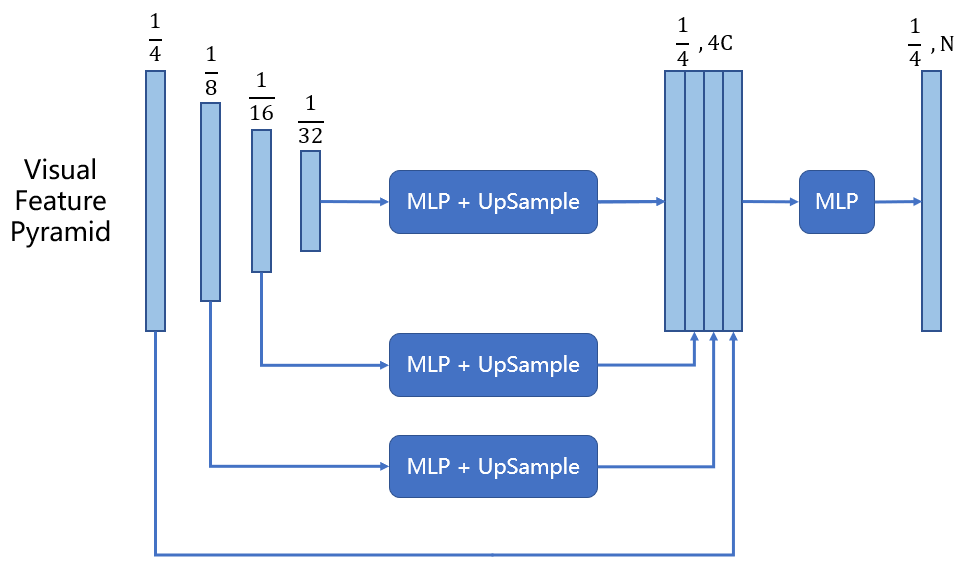}
         \caption{Dense decoder.}
         \label{fig:dense_decoder}
     \end{subfigure}
     \hfill
     \begin{subfigure}[b]{0.9\linewidth}
         \centering
         \includegraphics[width=\linewidth]{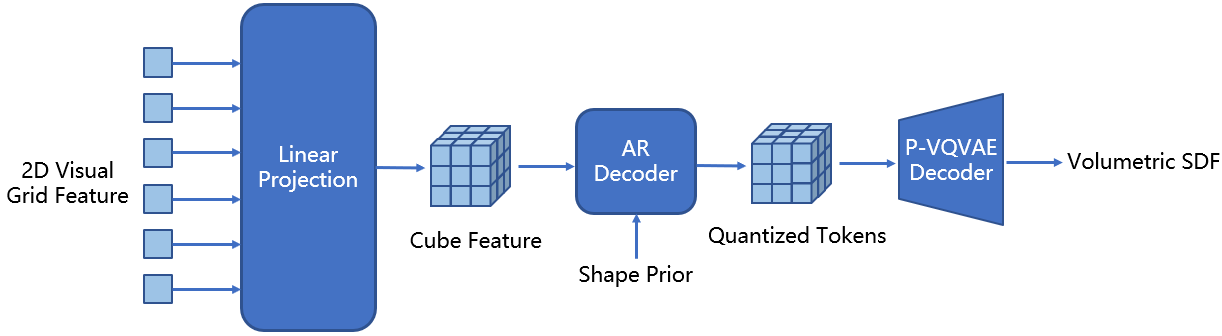}
         \caption{3D decoder.}
         \label{fig:3d_decoder}
     \end{subfigure}
     \vspace{5pt}
    \caption{An illustration of the decoders. Let $N$ denote the output dimension, \eg 4 for the bounding box. (a) The output of Label decoder is a single vector of length $N$. (b) The output of the Dense decoder is a dense grid with each element being an $N$-length vector. This dense grid is then upsampled back to the original image dimension through bilinear interpolation. (c) The output of the 3D decoder is a volumetric SDF. The Autoregressive(AR) decoder is a fixed module with pre-trained weights from \citet{mittal2022autosdf}.}
    \label{fig:decoder}
\end{figure}

\begin{itemize}[leftmargin=*,noitemsep,nolistsep]
    \item \textbf{Label decoder.} We implement the label decoder as a stack of Transformer blocks. The decoder flattens the last feature map from the encoder and then regards the feature as a sequence of tokens, to which fixed 2D positional embedding is applied. We concatenate the text token sequence with the visual feature sequence if needed. We insert an extra label token as the proxy to regress desired outputs. We pass this token through an MLP after Transformer layers for the final predictions. Note that we do not adopt this decoder in the image-text retrieval task, where we instead model the cosine similarity of embeddings right after simple linear layers.
    \item \textbf{Dense decoder.} We adopt the lightweight Segformer head \citep{xie2021segformer} as our dense decoder to handle dense prediction tasks. It performs efficient spatial and channel-wise feature aggregation on the visual feature pyramid representation.
    \item \textbf{3D decoder.} Implemented for the single-view 3D reconstruction task, the 3D decoder consists of three sub-decoders. The first up-convolution decoder maps 2D visual representation into a conditional distribution over the latent variables of 3D shapes. We then combine the predicted conditional distribution with autoregressive priors produced by a pre-trained Transformer-based decoder \citep{mittal2022autosdf} to obtain a joint distribution over 3D shapes. We finally obtain the volumetric SDF by forwarding the joint distribution to the third decoder, the decoder of a pre-trained Patch-wise Encoding VQ-VAE \citep{mittal2022autosdf}. We freeze the weights except for the first sub-decoder.
    \item \textbf{Navigation}. Following \citet{hong2021vln}, we use visual backbones to pre-compute visual representations (using the last feature map) at every possible viewpoint in a given room setting to accelerate RL training. We then adopt Recurrent VLN Bert as the task decoder, specifically the PREVALENT-like \citep{hao2020prevalent} variant due to its superior performance over the original OSCAR-like \citep{li2020oscar} model. And as in that paper, we initialize the decoder with pre-trained PREVALENT weights and use a combination of RL loss and imitation learning (IL) loss as the training objective.
    \item \textbf{Manipulation}. This decoder is inherited from CLIPort \citep{shridhar2022cliport}. Similar to the dense decoder, this decoder depends on the feature pyramid from visual representation and performs up-sampling at each level. Differently, language instructions are required to finish the manipulation task. To process language input, the decoder tiles text features and aggregates them with visual features. The output of the decoder is a dense affordance map to guide the agent in picking and placing. It is important to bring up that we remove the extra spatial stream to prevent the intervention of other visual representations, while the original CLIPort \citep{shridhar2022cliport} includes that stream to attain improved performance. 
\end{itemize}

\begin{table}[ht!]
    \centering
    \caption{Settings for \textit{Perceive}, \textit{Ground}, and \textit{Reason} tasks. \textit{Act} tasks are omitted here since they belong to separate training schemes. Abbreviations: ``Cam. Pose.'' for camera pose estimation, ``I-T Retr.'' for image-to-text retrieval, ``Phr. Grnd.'' for phrase grounding, ``Sem. Seg.'' for semantic segmentation, ``Com. Res'' for common sense reasoning, ``Abs. Res.'' for abstract reasoning.}
    \vskip 0.05in
    \label{tab:setting}
    \resizebox{\linewidth}{!}{%
        \begin{tabular}{cccccccccc}
        \toprule
        Setting & Depth Est. & Cam. Pose. & 3D Recon. & I-T Retr. & Phr. Grnd. & Sem. Seg. & VQA & Com. Res & Abs. Res. \\
        \midrule
        Decoder & Dense & Label & 3D & Linear & Label & Dense & Label & Label & Label \\
        Out Dim & 1 & 7 & 512 & 1024 & 4 & 151 & 1843 & 4 & 2 \\
        Batch Size & 128 & 32 & 32 & 256 & 128 & 128 & 128 & 128 & 32\\
        Epoch & 50 & 100 & 100 & 50 & 50 & 50 & 30 & 20 & 30 \\
        \bottomrule
        \end{tabular}
    }%
\end{table}

\paragraph{Training.} Most of the training settings are listed in \cref{tab:setting}. We use AdamW \citep{loshchilov2017decoupled} optimizer, with $\beta_1=0.9,\beta_2=0.99$ and weight decay at $10^{-2}$. During training, we warm up the learning rate linearly in the first 2\% steps to reach $1\times10^{-4}$ and adopt a linear decay schedule gradually decreasing the learning rate to zero during the rest of steps. The gradient norm is clipped to 1. Navigation and manipulation tasks are trained in simulation environments and stopped when convergence is observed. For simplicity, we resize input images to 224 to align with the resolution during the pre-training phase of visual representations, which is necessary since these representations stay fixed during the evaluation phase. All experiments are affordable on a single A100 GPU.

\paragraph{Summary score.} To summarize the overall performance of a visual representation on various tasks and conclude how general-purpose it is, we propose a comprehensive summary score. To obtain such a score, we need to unify divergent metrics among the tasks. In general, there are two types of metrics in our benchmark: percentage and error. We make a normalized mapping to convert all error metrics to percentage metrics: given error $E$, the corresponding metric in percentage form $P$ is formulated as $P(E)=e^{-1.386E}$, where the constant $-1.386$ is determined by two conditions: $P(0)=1$ and $P(0.5)=0.5$. With this normalization, we can calculate a score for each task by averaging all aligned metrics within the task, and then calculate the final summary score for a visual representation by averaging scores on all 11 tasks. To make domains more balanced, we augment each task in \textit{Act} with a weight at $1.5$.

\section{Ablation}\label{sec:ablation}
\subsection{Finetuning}
In the experiment section, we mainly report results of fixed visual representations, since the evaluation of general-purpose requires a shared visual representation for various tasks. To further study the effects of finetuning the visual backbones, we provide experiments on two visual representations learned in a self-supervised fashion: RN-MoCo and ViT-16-MAE. During finetuning, we set the base learning rate of visual backbones to $10^{-5}$ and keep the same warm-up and linear decay schedule. As shown in \cref{tab:finetune}, dense prediction tasks, \eg depth estimation, benefit most from finetuning visual backbones. Other tasks in general can benefit marginally from finetuning. While a few cases turn out to be harmful finetuning. We hypothesize that learning the decoder with a dynamic encoder is hard and prone to overfitting, especially on a relatively small amount of training data.

{
\newcommand{\sepline}{\hline}
\renewcommand{\arraystretch}{1.2}
\begin{table}[t!]

\centering
\caption{Influence of finetuning visual backbones. For space considerations, we use the abbreviation of words for identifying tasks (\eg ``Cam. Pose.'' for camera pose estimation, ``I-T Retr.'' for image-to-text retrieval, ``Phr. Grnd.'' for phrase grounding, ``Sem. Seg.'' for semantic segmentation, ``Com. Res'' for common sense reasoning, ``Abs. Res.'' for abstract reasoning). Finetuning on Act tasks is omitted here. We also remark on the gains from finetuning in a separate column. \textcolor{Green}{Green} indicates gains, while \textcolor{Red}{red} indicates drops.}
\label{tab:finetune}
\vskip 0.2in

\resizebox{.9\linewidth}{!}{
\begin{tabular}{c|c|ccc|ccc}
\toprule
\multicolumn{2}{c}{\multirow{2}[2]{*}{\textbf{Task}}} & \multicolumn{3}{c}{RN-MoCo} & \multicolumn{3}{c}{ViT-16-MAE} \\
\cmidrule(lr){3-5}\cmidrule(lr){6-8}
\multicolumn{2}{c}{} & Fixed & Finetune & \multicolumn{1}{c}{Gain} & Fixed & Finetune & Gain \\
\hline
\multicolumn{2}{c}{\textbf{Perceive}} & \multicolumn{3}{c}{} & & & \\
\hline
\multirow{3}{*}{Depth Est.} & test d\textless{}1.25 $\uparrow$ & 0.6400 & 0.7302 & \textcolor{Green}{+0.0902} & 0.7699 & 0.8605 & \textcolor{Green}{+0.0906} \\
 & test AbsRel $\downarrow$ & 0.2184 & 0.1725 & \textcolor{Green}{-0.0459} & 0.1554 & 0.1203 & \textcolor{Green}{-0.0351} \\
 & test RMSE $\downarrow$ & 0.7284 & 0.6042 & \textcolor{Green}{-0.1242} & 0.5486 & 0.4444 & \textcolor{Green}{-0.1042} \\
 \cline{2-8}
 \multirow{4}{*}{Cam. Pose} & Trans.(CL) $\downarrow$& 1.972 & 1.943 & \textcolor{Green}{-0.029} & 2.177 & 1.926 & \textcolor{Green}{-0.251} \\
 & Orient.(CL) $\downarrow$& 5.427 & 5.671 & \textcolor{Red}{+0.244} & 4.911 & 3.837 & \textcolor{Green}{-1.074} \\
 & Trans.(7S) $\downarrow$& 0.241 & 0.255 & \textcolor{Red}{+0.014} & 0.271 & 0.250 & \textcolor{Green}{-0.021} \\
 & Orient.(7S) $\downarrow$ & 8.797 & 8.452 & \textcolor{Green}{-0.345} & 6.699 & 5.823 & \textcolor{Green}{-0.876} \\
 \cline{2-8}
 3D Recon. & test mIoU $\uparrow$& 41.98 & 42.87 & \textcolor{Green}{+0.89} & 43.95 & 48.25 & \textcolor{Green}{+4.30} \\
 \sepline
\multicolumn{2}{c}{\textbf{Ground}} & \multicolumn{3}{c}{} & & & \\
\hline
 \multirow{3}{*}{I-T Retr.} & test R@1 $\uparrow$& 23.3 & 18.1 & \textcolor{Red}{-5.2} & 23.4 & 32.1 & \textcolor{Green}{+8.7} \\
 & test R@5 $\uparrow$& 48.8 & 42.4 & \textcolor{Red}{-6.4} & 50.5 & 59.2 & \textcolor{Green}{+8.7} \\
 & test R@10 $\uparrow$& 60.0 & 53.7 & \textcolor{Red}{-6.3} & 63.6 & 70.3 & \textcolor{Green}{+6.7} \\
  \cline{2-8}
 \multirow{3}{*}{Phr. Grnd.} & val Acc@0.5 $\uparrow$& 54.40 & 56.73 & \textcolor{Green}{+2.33} & 65.18 & 64.55 & \textcolor{Red}{-0.63} \\
 & testA Acc@0.5 $\uparrow$& 59.14 & 61.37 & \textcolor{Green}{+2.23} & 67.59 & 66.13 & \textcolor{Red}{-1.46} \\
 & testB Acc@0.5 $\uparrow$& 50.88 & 52.10 & \textcolor{Green}{+1.22} & 62.10 & 62.38 & \textcolor{Green}{+0.28} \\
 \cline{2-8}
 Sem. Seg. & val mIoU $\uparrow$& 18.05 & 27.40 & \textcolor{Green}{+9.35} & 26.25 & 35.21 & \textcolor{Green}{+8.96} \\
\sepline
\multicolumn{2}{c}{\textbf{Reason}} & \multicolumn{3}{c}{} & & & \\
\hline
\multirow{2}{*}{VQA} & val Acc. $\uparrow$ & 50.20 & 51.49 & \textcolor{Green}{+1.29} & 54.82 & 54.49 & \textcolor{Red}{-0.33} \\
 & test-dev Acc. $\uparrow$ & 45.02 & 45.44 & \textcolor{Green}{+0.42} & 48.50 & 47.89 & \textcolor{Red}{-0.61} \\
 \cline{2-8}
 Com. Res. & val Acc. $\uparrow$ & 50.98 & 55.22 & \textcolor{Green}{+4.24} & 60.76 & 55.68 & \textcolor{Red}{-5.08} \\
 \cline{2-8}
 \multirow{2}{*}{Abs. Res.} & val Acc. $\uparrow$ & 62.38 & 62.58 & \textcolor{Green}{+0.20} & 62.12 & 64.38 & \textcolor{Green}{+2.26} \\
 & test Acc. $\uparrow$ & 64.63 & 62.36 & \textcolor{Red}{-2.27} & 62.85 & 65.14 & \textcolor{Green}{+2.29} \\
\bottomrule
\end{tabular}
}

\end{table}
}

\subsection{Resolution}
Another factor that may cause inferior performance is the resolution of the input image. We demonstrate this by showing performance gain from scaling up input resolution, see \cref{tab:highreso}. The results reveal the headroom of performance when scaling up resolution.
\begin{table}[ht!]
    \centering
    \caption{Performance progressively gains from fine-tuning and higher resolution. Take RN-IN on phrase grounding task for example. The results are reported in Acc@0.5 metrics.}
    \vskip 0.05in
    \label{tab:highreso}
    \resizebox{0.38\linewidth}{!}{%
        \begin{tabular}{lccc}
        \toprule
         & val & testA & testB \\
        \midrule
        Fixed on 224 & 48.57 & 54.48 & 43.87 \\
        Tuned on 224 & 58.65 & 64.15 & 53.26 \\
        Tuned on 512 & 65.55 & 71.95 & 59.83 \\
        \bottomrule
        \end{tabular}
    }%
\end{table}

\subsection{Language Encoder}
\paragraph{Image-text retrieval.} As mentioned before, we observe unsatisfactory performances on the image-text retrieval task from the experimental results in \cref{tab:all_results}. We hypothesize this partially stems from the language encoder. Initially, we select RoBERTa as the default language encoder to provide disentangled language representation regardless of the visual branch, which can make the comparison fair. Moreover, RoBERTa has no bias towards specific visual representation due to the text-only pre-training. Nevertheless, such a setting poses more challenges for the visual encoder since the language representation lacks alignment with the visual domain. To examine this issue, we use the CLIP language encoder \citep{radford2021learning} to perform the image-text retrieval task for comparison. \cref{tab:retrieval} shows that switching to the CLIP language encoder leads to significant improvement. This demonstrates the effectiveness of coupled representations and explains the lower performance when using RoBERTa.

\begin{table}[ht!]
    \centering
    \caption{Performance of different vision-language encoder pairs on Flickr30k test set.}
    \vskip 0.05in
    \label{tab:retrieval}
    \resizebox{0.6\linewidth}{!}{%
        \begin{tabular}{lccc}
        \toprule
        Encoder Pair (V+L) & Recall@1 & Recall@5 & Recall@10 \\
        \midrule
        ViT-16-CLIP + RoBERTa & 59.5 & 86.0 & 92.5 \\
        ViT-16-CLIP + GPT2-CLIP & 81.6 & 96.9 & 98.7 \\
        \bottomrule
        \end{tabular}
    }%
\end{table}

To address the gap between disentangled visual encoder and language encoder, we add adapters following the representations produced by the two encoders. On the other hand, we would not hope the adapters to be heavy, because learning these extra parameters from scratch requires extensive multi-modal training and is prone to overfitting. This is confirmed by our observation that adding heavier adapters would lead to a performance drop. Consequently, we choose to limit the computation for modality fusion down to linear projection. Such a design would have the retrieval highly rely on the merit of pre-trained representations, which makes the evaluation strict for visual representations on this task.

As far as we know, the brilliant performance of current state-of-the-art models on the image-text retrieval task mostly credits to large-scale vision-language pre-training on coupled representations. Besides, fixed language encoder initialing from text-only pre-training proves sub-optimal results on vision-language tasks, as reported in \citep{kim2021vilt}. To this end, how to evaluate visual representations under a language-agnostic and small-data regime is cast as a challenging yet unsolved problem.


\paragraph{Navigation.} Similar to Image-text retrieval, we also conduct ablation of language encoder on navigation task. It is customary in the vision-language navigation (VLN) task to use a pre-trained language encoder that produces visually-aligned representations to better aid navigation. And we are also interested in how much such a gap can affect the performance on navigation. We re-evaluate the visual representations with the more VLN-fitting LXMERT~\citep{tan2019lxmert} language encoder from \citet{hong2021vln} apart from the results using RoBERTa. We show a full comparison between LXMERT and RoBERTa as the language encoder in \cref{tab:navigation}.

\begin{table}[ht!]
    \centering
    \caption{Exhaustive ablation on language encoder (RoBERTa \vs LXMERT) on navigation task. ResNet shows stronger fitting on seen scenarios, while ViT generalizes better on unseen scenarios.}
    \vskip 0.05in
    \label{tab:navigation}
    \resizebox{0.8\linewidth}{!}{%
        \begin{tabular}{lllll}
        \toprule
        Encoder Pair (V+L) & Seen Succ. & Seen SPL & Unseen Succ. & Unseen SPL  \\
        \midrule
        RN-IN + RoBERTa & \underline{\textbf{54.75}} & \underline{\textbf{51.02}} & 48.62 & 42.61 \\
        \hphantom{RN-IN} + LXMERT & 58.18 \textcolor{Green}{(+3.43)} & 54.28 \textcolor{Green}{(+3.26)} & 55.17 \textcolor{Green}{(+6.55)} & 50.40 \textcolor{Green}{(+7.79)}\\
        RN-MoCo + RoBERTa & 52.50 & 48.92 & 48.49 & 43.49 \\
        \hphantom{RN-MoCo} + LXMERT & 57.49 \textcolor{Green}{(+4.99)} & 54.20 \textcolor{Green}{(+5.28)} & 55.21 \textcolor{Green}{(+6.72)} & 50.73 \textcolor{Green}{(+7.24)}\\
        RN-CLIP + RoBERTa & 52.40 & 48.39 & 49.00 & 43.49 \\
        \hphantom{RN-CLIP} + LXMERT & 58.86 \textcolor{Green}{(+6.46)} & 55.03 \textcolor{Green}{(+6.64)} & 55.98  \textcolor{Green}{(+6.98)} & 50.73 \textcolor{Green}{(+7.24)} \\
        RN-Ego + RoBERTa & 44.07 & 39.41 & 43.89 & 37.88 \\
        \hphantom{RN-Ego} + LXMERT & 45.45 \textcolor{Green}{(+1.38)} & 41.89 \textcolor{Green}{(+2.48)} & 48.45 \textcolor{Green}{(+4.56)} & 44.00 \textcolor{Green}{(+6.12)} \\
        ViT-32-CLIP + RoBERTa & 51.03 & 46.98 & 49.21 & \underline{\textbf{43.99}} \\
        \hphantom{ViT-32-CLIP} + LXMERT & 54.95 \textcolor{Green}{(+3.92)} & 50.53 \textcolor{Green}{(+3.55)} & 56.88 \textcolor{Green}{(+7.67)} & 51.70 \textcolor{Green}{(+7.71)}\\
        ViT-16-CLIP + RoBERTa & 46.23 & 44.15 & 47.13 & 43.60 \\
        \hphantom{ViT-16-CLIP} + LXMERT & \underline{\textbf{61.90}} \underline{\textbf{\textcolor{Green}{(+15.67)}}} & \underline{\textbf{55.22}} \underline{\textbf{\textcolor{Green}{(+11.07)}}} & \underline{\textbf{59.51}} \underline{\textbf{\textcolor{Green}{(+12.38)}}} & \underline{\textbf{53.15}} \underline{\textbf{\textcolor{Green}{(+9.55)}}} \\
        ViT-16-MAE + RoBERTa & 48.87 & 43.91 & \underline{\textbf{49.55}} & 43.72 \\
        \hphantom{ViT-16-MAE} + LXMERT & 48.29 \textcolor{Red}{(-0.58)} & 43.16 \textcolor{Red}{(-0.75)} & 52.41 \textcolor{Green}{(+2.86)} & 45.99 \textcolor{Green}{(+2.27)}\\
        \bottomrule
        \end{tabular}
    }
\end{table}

We observe that performances improve on nearly all visual representations when we use LXMERT instead of the text-only pre-trained RoBERTa. This demonstrates that visually-aligned language representations do massively improve downstream performances. And to further eliminate the influence of the subsequent navigation decoder pre-trained along with the LXMERT language encoder, we conduct an extra ablation to test whether a from-scratch navigation decoder with randomly initialized weights, which loses the alignment from pre-trained coupled encoder-decoder, can still achieve similar results. The results are shown in \cref{tab:navfromscratch}.

\begin{table}[ht!]
    \centering
    \caption{Performance of LXMERT language encoder paired with pre-trained decoder \vs from-scratch decoder.}
    \vskip 0.05in
    \label{tab:navfromscratch}
    \resizebox{0.8\linewidth}{!}{%
        \begin{tabular}{llllll}
        \toprule
        Encoder Pair (V+L) & Decoder & Seen Succ. & Seen SPL & Unseen Succ. & Unseen SPL  \\
        \midrule
        ViT-CLIP-16 + LXMERT & pre-trained & 61.90 & 55.22 & 59.51 & 53.15 \\
        ViT-CLIP-16 + LXMERT & from-scratch & 60.92 \textcolor{Red}{(-0.98)} & 56.54 \textcolor{Green}{(+1.32)} & 57.13 \textcolor{Red}{(-2.38)} & 51.40 \textcolor{Red}{(-1.75)} \\
        \bottomrule
        \end{tabular}
    }%
\end{table}

We observe that the model with a from-scratch navigation decoder still achieves comparable performance despite losing the encoder-decoder pre-training alignment. This confirms that the improvements shown in \cref{tab:navigation} are not due to the encoder-decoder coupling, but rather due to the visually-aligned semantics of the LXMERT language encoder.


\section{Social Impact}
 With the emergence of foundation models, tremendous efforts have been put into improving the multi-tasking capabilities of large-scale pre-trained models. Instead of improvements from the model's side, we focus on the problem of designing a general, sophisticated and fair evaluation metric that reveals the models' capabilities on the full spectrum of vision tasks. We believe our G-VUE benchmark will make an excellent complement to existing works as they can be easily adapted to be evaluated in G-VUE. There is no known negative societal impact at this point.

\section{License}
We list the licenses of all datasets involved in G-VUE as follows:
\begin{itemize}
    \item NYUv2: Unknown
    \item ShapeNet: \href{https://shapenet.org/terms}{Custom}
    \item CambridgeLandmarks: \href{http://creativecommons.org/licenses/by-nc-sa/2.0/uk/}{CC BY-NC-SA 2.0 UK}
    \item 7-Scenes: \href{https://www.microsoft.com/en-us/research/wp-content/uploads/2016/02/7-scenes-msr-la-dataset-7-scenes.rtf}{Custom}
    \item Flickr30k: \href{https://www.flickr.com/help/terms/}{Custom}
    \item RefCOCO: \href{https://opensource.org/licenses/Apache-2.0}{Apache License 2.0}
    \item ADE20k: \href{https://opensource.org/licenses/BSD-3-Clause}{3-Clause BSD License}
    \item GQA: \href{https://creativecommons.org/licenses/by/4.0/}{CC BY 4.0}
    \item VCR: \href{https://visualcommonsense.com/license/}{Custom}
    \item Bongard-HOI: \href{https://github.com/NVlabs/Bongard-HOI/blob/master/LICENSE}{Custom}
    \item R2R: Unknown
    \item Ravens: \href{https://opensource.org/licenses/Apache-2.0}{Apache License 2.0}
\end{itemize}


\end{document}